\documentclass[journal,11pt,singlespace,twocolumn]{IEEEtran}

\usepackage{color,soul}
\usepackage{amsmath}
\usepackage{graphicx,psfrag,epsf}
\usepackage{enumerate}
\usepackage{url}
\usepackage{hyperref}
\usepackage{amssymb,amsfonts,mathtools}
\usepackage{bm}
\usepackage{color, colortbl}
\usepackage{algorithm}
\usepackage{algorithmic}
\usepackage{balance,cite}
\usepackage{multirow}
\usepackage{wrapfig,bbm}

\DeclareMathOperator*{\argmin}{arg\,min}
\DeclareMathOperator*{\argmax}{arg\,max}
\DeclarePairedDelimiterX{\norm}[1]{\lVert}{\rVert}{#1}
\DeclarePairedDelimiterX{\bnorm}[1]{\biggl\lVert}{\biggr\rVert}{#1}
\DeclarePairedDelimiterX{\abs}[1]{\lvert}{\rvert}{#1}

\newtheorem{proposition}{Proposition}

\def\T{{ \mathrm{\scriptscriptstyle T} }} %transpose
\def\A{{\texttt{A}}}
\def\B{{\texttt{B}}}
\def\F{\mathcal{F}}
\def\Xb{\mathbf{X}}
\def\Yb{\mathbf{Y}}
\def\xb{\mathbf{x}}
\def\yb{\mathbf{y}}
\def\wb{\mathbf{w}}

\def\X{\mathcal{X}}
\def\Y{\mathcal{Y}}
\def\bm{\mathbf}
\def\R{\mathbb{R}}
\def\gb{\mathbf{g}}
\def\fb{\mathbf{f}}

\usepackage{xcolor}
\colorlet{linkequation}{blue}
\newcommand*{\eqrefblue}[1]{%
  \begingroup
    \hypersetup{
      linkcolor=linkequation,
      linkbordercolor=linkequation,
    }%
    (\ref{#1})%
  \endgroup
}

\title{ASCII: ASsisted Classification with \\ Ignorance Interchange}
\author{Jiaying~Zhou\textsuperscript{\textdagger},
        Xun~Xian\textsuperscript{\textdagger},
        Na~Li\textsuperscript{\textasteriskcentered},
        and  Jie~Ding\textsuperscript{\textdagger}
\thanks{\textsuperscript{\textdagger} J.~Zhou, X.~Xian, and J.~Ding are with the School of Statistics, University of Minnesota.
\textsuperscript{\textasteriskcentered} N.~Li is with the School of Engineering and Applied Sciences, Harvard University.
}
\thanks{This research was funded by the Army Research Office (ARO) under grant number W911NF-20-1-0222, and the GIA award from the Office of the Vice President for Research, University of Minnesota.}
}

\date{\today}

\begin{document} 
\maketitle
\begin{abstract}
The rapid development in data collecting devices and computation platforms produces an emerging number of agents, each equipped with a unique data modality over a particular population of subjects. While an agent’s predictive performance may be enhanced by transmitting others’ data to it, this is often unrealistic due to intractable transmission costs and security concerns. In this paper, we propose a method named ASCII for an agent to improve its classification performance through assistance from other agents. The main idea is to iteratively interchange an ignorance value between 0 and 1 for each collated sample among agents, where the value represents the urgency of further assistance needed. The method is naturally suitable for privacy-aware, transmission-economical, and decentralized learning scenarios. The method is also general as it allows the agents to use arbitrary classifiers such as logistic regression, ensemble tree, and neural network, and they may be heterogeneous among agents. We demonstrate the proposed method with extensive experimental studies.

\end{abstract}


\section{Introduction} \label{sec_intro}
%\vspace{-0.2cm}
%============================================================================
%============================================================================
%In recent years, there has been a growing interest in statistical learning problems with a set of decentralized agents, where each agent encompasses a specific data modality and a statistical model built using domain-specific knowledge.

In recent years, the advancement of mobile devices and information technology has led to an emerging number of distributed multimodal data sources for different populations, e.g., a group of patients and a cohort of mobile users~\cite{xian2020assisted}. %a basket of commercial products, etc. 
Each source of data is often collected and held by an agent with domain-specific interest. 
For example, suppose that agent $\A$ is a grocery store that aims to predict customers’ shopping preferences using its shopping records, and agent $\B$ is an IT service company with the customers’ mobile data.
Both $\A$ and $\B$ hold a unique set of features from the same population. In other words, they hold two submatrices of a holistic data matrix (in hindsight) where rows are indexed by customer IDs. 
%$\A$ and $\B$ hold heterogeneous models 
Intuitively, if Agent $\A$ can build a model based on the holistic data matrix, the prediction performances would typically be better than model trained only with $\A$'s data. 

However, constructing such a holistic matrix may give rise to certain problems, e.g., breach of data privacy~\cite{yao1982protocols,chaum1988multiparty,dwork2004privacy,gentry2009fully,dwork2011differential,armknecht2015guide},  infeasible transmission cost~\cite{stojanovic2009underwater,schettini2010underwater}, and hardware capacity~\cite{DingDrasic,DingController}. Moreover, data fusion often assumes a centralized dataset held by one agent~\cite{lahat2015multimodal}. 
%Hypothetically, all the data may be collated (e.g., by timestamps or data IDs) into a larger database, providing holistic information set for the underlying subjects. 
%
%Such a hypothetical dataset may be helpful for each agent to fully extract useful information.
%However, it is often unrealistic to collate multiple data sources/modalities into one dataset due to various 
%concerns such as data privacy~\cite{yao1982protocols,chaum1988multiparty,dwork2004privacy,gentry2009fully,dwork2011differential,armknecht2015guide}, limited bandwidth~\cite{stojanovic2009underwater,schettini2010underwater,DingDrasic}, and hardware capacity...
%Moreover, an agent may not wish to publicize its learning task and learning model due to ethical or business concerns.
The above challenges motivate the following question. \textit{Can any particular agent improve its learning quality with the assistance of other agents, but without transmitting their private data or models?}
In this paper, we present a general solution named \underline{\textbf{AS}}sisted \underline{\textbf{C}}lassification with \underline{\textbf{I}}gnorance \underline{\textbf{I}}nterchange (ASCII), to  address the above challenge. 
The ASCII allows each agent to assist other agents by interchanging communication-efficient statistics instead of raw data. Consequently, such assistance enables a significantly better performance compared with single-agent learning (without assistance). Such assistance will also allow agents to use their own private models (or algorithms) and data modality.

A related work is \textit{Federated Learning} (FL)~\cite{shokri2015privacy,konevcny2016federated,mcmahan2016communication}, which is a distributed learning framework that features communication efficiency. 
The main idea of FL to learn a joint model using the averaging of locally learned model parameters, so that the training data do not need to be transmitted. 
To address heterogeneous data, there exist some recent work of vertically-partitioned FL, including those based on the homomorphic encryption and partial stochastic gradient descent~\cite{cheng2019secureboost, fengy2019securegbm, yang2019federated,liu2019communication}.
While FL agents are required to use a commonly shared global model, our assisted learning allows agents to autonomously use their own (private) models as well as data. 

 % which may be appealing in some model-private cooperative learning scenarios.

The main idea of our method is described as follows. 
First, we let $\A$ fit its model, and then evaluate the model on each sample to obtain an `ignorance' score, which is a value between 0 and 1 indicating the extent to which a sample may not be ignored. 
A substantial ignorance score (say 1) means that the corresponding sample has not been adequately modeled, and thus extra information is needed from other agents. % for improvement.
Second, $\A$ sends the labels and ignorance scores to $\B$, who uses its data and model to train a classifier using weighted samples. Here, $\B$'s sample weights will be the current ignorance scores so that $\B$ can focus on those data that have not been well-modeled before. After that, $\B$ updates the ignorance scores and sends them back to $\A$, who will then initialize the second round of information interchange. The above interactions are repeated until a stop criterion is met.
For future inference, $\B$ will apply an ensemble model created from its local models (when interacting with $\A$) to its new data observation and sends the prediction result to $\A$. 
$\A$ will then integrate it with its prediction result to make the final decision. 
In this way, $\B$ provides side information to $\A$  to eventually improve $\A$'s learning task. 
Note that the above learning mechanism only needs $\A$ and $\B$ to exchange numerical labels (for local training), data IDs (for data collation), and ignorance scores (for side information), instead of raw data. Moreover, agents are free to use their own learning models.
%The class labels of people are in public, while different features are stored privately in each data holder. When there are $M$ agents, named $\A$, $\B$,\ldots, $\A$ wants to get some complementary information from other agents that might be helpful in classification, and the agents tend to assist $\A$ without transmitting original data, or even reasonably perturbed data, under the concern of intractable transmission cost. 
In the above procedure, the ignorance scores are mathematically derived so that the final classifier of $\A$ as assisted from $\B$ will approximate the oracle classifier using the hypothetically collated data from $\A$ and $\B$.

% Similarity and difference with AdaBoost

Our derivation of the algorithm was inspired by the pioneering work of Adaptive Boosting (AdaBoost)~\cite{freund1995desicion,margineantu1997pruning,friedman2000additive,hastie2009multi}, where weak learners are sequentially created and aggregated into a strong learner. 
Like AdaBoost, our approach will also create a sequence of models trained from weighted samples. %an ensemble of models will be created to assist an agent,
The main difference between AdaBoost and our method is two folds. First, AdaBoost performs training on all the data, while our approach is based on training heterogeneous data held by different agents. Second, in our approach, each agent's sample weights are calculated from both the agent's previous sample weights and other agents' transmitted weights (`ignorance').
Our method may be regarded as a generalization of AdaBoost to address privacy-aware distributed learning with vertically-split variables.

%Conceptually, our ignorance scores and local models at each round of assistance are the counterparts of sample weights and weak agents in the context of AdaBoost. 
%The main difference is that the training cannot be performed on all the data. The information exchanged by $\A$ and $\B$ needs to ensure a proper learning  objective of $\A$. Clearly, the learning efficacy from our procedure should be better than a vanilla learning approach using only $\A$'s or $\B$'s data and is preferably near the oracle achieved by pulling all the data together.  

%\textbf{Main contributions}.
The main contributions of this work are three folds. First, we propose an assisted learning method where an agent can significantly improve its learning performance by interchanging side information with other agents, without transmitting private data or private models. 
Second, we establish the method in a general `model-free' framework, meaning that we allow each agent to employ its own model. As a result, agents do not necessarily use the same global model
%The method does not need a prescribed global model for all the agents. 
or a trusted third party for coordination, appealing in many autonomous learning scenarios. % [...].
Third, we develop theoretical justifications and interpretations of the derived ignorance scores.
We show by extensive experiments that the proposed solution will significantly enhance single-agent learning performance. Moreover, in many cases, the method also produces near-oracle performance, where the oracle is defined as the performance from hypothetically collated dataset.
% We provide various  studies to demonstrate the near-oracle performance of the proposed method.

%\textbf{Outline}.
The outline of the paper is given below.
In Section~\ref{sec_back}, we introduce some notation and formulate the problem.
In Section~\ref{sec_AL2}, we propose a general approach for assisted classification in a two-agent scenario and discuss its theoretical justifications.
In Section~\ref{sec_ALm}, we extend the proposed method to a multi-agent scenario.
In Section~\ref{sec_other}, we introduce some variants of the proposed algorithm, and highlight the unique advantages of our proposal in later numerical comparisons. 
%In Section~\ref{sec_otherview}, we provide an interpretation and justification of the proposed approach from another perspective. 
%In Section~\ref{sec_thm3}, we propose the notation of virtual database, and propose an algorithm to learn in a general multi-agent scenario.
We provide extensive experimental studies in Section~\ref{sec_exp}, and conclude the work in Section~\ref{sec_con}.

%\subsection{Related work}
% Our work is related with the following two areas.
% \textit{Data Fusion}~\cite{ngiam2011multimodal,srivastava2012multimodal, lahat2015multimodal,baltruvsaitis2018multimodal}, also known as multimodal learning, aims to aggregate information from multiple modalities to perform prediction. 
% % For example, recent data fusion applications of using deep networks to learn features over different modalities~\cite{srivastava2012multimodal,ngiam2011multimodal}. 
% % The main focus is to efficiently and effectively extract and integrate complementary information.   %because simply fusing multiple datasets may not give the most useful results~\cite{lahat2015multimodal}. 
% While data fusion  our work aims to construct a assisted learning framework without sharing  agent's training data and models.  %under \textit{privacy}. 
%Moreover, an agent may not wish to publicize its learning task and learning model due to ethical or business concerns.

% \textit{Transfer Learning}~\cite{pan2009survey,dai2007boosting} aims to improve the predictive performance using training data whose distribution is possibly different from the testing data. 
% % while in our setting, agents hold heterogeneous features from the same objects, 
% While transfer learning's setting is on heterogeneous domains, which is different from ours that we focus on vertical-split data for same subjects. 

%\vspace{-0.2cm}
\section{Problem} \label{sec_back}
%\vspace{-0.3cm}

\subsection{Background and Notation} \label{subsec_notation}
%\vspace{-0.3cm}

%We now give some notation regarding our problem. Similar as the notation in  \cite{hastie2009multi}, 
We first introduce some notation. 
Suppose there exist $M$ agents, and the $m$-th agent holds $\mathbf X^{(m)} = [\mathbf x_1^{(m)},\ldots,\mathbf x_{n}^{(m)}]^\T \in \mathbb R^{n\times p_m}$ as the private data matrix, where $n$ is the sample size and $p_m$ is the number of feature variables, $m=1,\ldots, M$.
We will consider a classification problem that involves $K$ classes ($K \geq 2$).
Let $\mathbf c = [c_1,c_2,\ldots,c_n]^\T $ denote the $K$-class label vector accessible by all the agents, where $c_i \in \{1,2,\ldots,K\}$.
Suppose that when the agents exchange information, they have a consensus on how to collate/align the data through a certain sample ID (e.g., person ID or timestamp). 
In the above setup, we implicitly assumed  that their data could be collated and features  are non-overlapping. %Without loss of generality
If the data are partially overlapping (in terms of sample ID), we suppose that only the overlapping data are used for technical convenience. %A more general setting as future work. 
%We denote $\mathcal F^{(k)} = \{f_1,f_2,\ldots,f_{p_k}\}$ as the feature sets in $\mathbf X^{(k)}$. 
To develop our technical approach, we will re-code the label vector $\mathbf c$ into a label matrix $\mathbf Y = [\mathbf y_1,\mathbf y_2,\ldots,\mathbf y_n]^\T$,
where each row $\mathbf y_i = [y_{i1},y_{i2},\ldots,y_{iK}]^\T$ with
\begin{equation}
	y_{ij} =\left\{
\begin{array}{cc}
 1    &   c_i = j\\
-\frac{1}{K-1} & c_i \neq j \\
\end{array}
\label{eq:y_space}
\right.
\end{equation}
 encodes the class $c_i$.
% The new label space $\mathcal Y$ is therefore defined as
% \begin{equation}
% \mathcal Y=\left\{
% \begin{array}{cccc}
% (1,-\frac{1}{K-1},-\frac{1}{K-1},\ldots,-\frac{1}{K-1}) \\
% (-\frac{1}{K-1},1,-\frac{1}{K-1},\ldots,-\frac{1}{K-1}) \\
% \vdots   \\
% (-\frac{1}{K-1},-\frac{1}{K-1},-\frac{1}{K-1},\ldots,1) \\
% \end{array}
% \right\}.
% \label{eq:labelspace}
% \end{equation}
Let $\X^{(m)}$ and $\Y$ denote the feature space of agent $m$ and label space, respectively.  
The main reason we use this encoding method is for technical convenience when implementing the exponential loss that we will elaborate in Section~\ref{subsec_adaboost}.
The above code format has been widely used, for multi-class classification tasks, e.g., %under sum-to-zero constraint, and it was first used in \cite{lee2004multicategory}
support vector machines~\cite{lee2004multicategory} and boosting~\cite{hastie2009multi}.

% \textcolor{red}{data collation}
% \textcolor{red}{focus on classification in this paper %, and cite the other two papers. 
% } 

%\vspace{-0.1cm}
\subsection{Problem Formulation} \label{subsec_adaboost} 
%\vspace{-0.1cm}
% \textcolor{red}{
% Introduce the Adaptive boosting algorithm and highlight the difference and new interpretations of the proposed algorithm 
% (refer to the pseudocode in Section~\ref{sec_AL2}). Also introduce what it means by weighted sample, and introduce   Algorithm~\ref{algo_supervise}.
% }
\subsubsection{Adaboost for the single-agent case}

Before we introduce the formulation for  multi-agent assisted learning, we briefly review the framework of single-agent learning and the AdaBoost algorithm for it.

Suppose that an agent is equipped with a data matrix $\bm X = [\bm  x_1,\ldots,\bm x_n]^\T$ and classification label matrix $\bm Y= [\bm y_1,\ldots,\bm y_n]^\T$. Note that $\bm y_i$ re-codes class $c_i$ into a length-$K$ vector.
The supervised learning task is to find a function $\bm f(\bm x) = [f_1(\bm x),f_2(\bm x),\ldots, f_K(\bm x)]^\T$ that approximates the function relationship between $\bm X$ and $\bm Y$. %to the multi-class case is expressed as
In many statistical learning contexts, the $\bm f$ is usually estimated by a risk minimization problem in the form of
\begin{equation}
\begin{aligned}
&	\min_{\bm f \in \F_T} \sum_{i=1}^n \ell( \bm y_i,  \bm f(\mathbf x_i))	
, %\quad 	\ell: (\mathbf y,\bm f) \mapsto e^{-\frac 1 K \mathbf y^\T \bm f},	
\\
&\text{subject to}\quad f_1(\bm x)+\ldots+f_K(\bm x) = 0,
\end{aligned}	
\label{opti}
\end{equation}
where $\ell$ and $\F_T$ are appropriately chosen loss function and model class, respectively. 
For a future (or testing) data $\tilde{\bm x}$, the learned model produces a prediction $\tilde{\yb}$ that corresponds to the prediction label $\tilde{c} = \argmax_i f_i(\tilde{\bm x})$.
Note that the above constraint is to ensure the identifiability of $\bm f$. % during  learning.

The AdaBoost algorithm~\cite{freund1995desicion,margineantu1997pruning,friedman2000additive,hastie2009multi} is derived based on a specific loss function and a model class.
In particular, it uses the following exponential loss function,
$$
\ell: (\mathbf y,\bm f) \mapsto e^{-\frac 1 K \mathbf y^\T \bm f},
$$
and the following model class (also referred to as `additive models'), \begin{align}
	\F_T = \biggl\{ \bm f:  \bm f(\mathbf x) = \sum_{t=1}^T \alpha_t \bm g_t(\mathbf x), \, \alpha_t \in \R,\,  \bm g_t \in \F_0 \biggr\}	. \nonumber % \label{eq_100}
\end{align}
where $T$ is the number of weak learners, and $\F_0$ is a class of basis functions mapping from $\X$ to $\Y$ (e.g., decision trees and linear classifiers), and $\alpha_t$ and $\bm g_t$, $t=1,\ldots,T$ are the unknown parameters/functions.

The vector form of $\bm f(\bm x)$ as derived by  \cite{hastie2009multi} implies that given $\bm x$, $\bm g_t(\bm x)$ maps $\bm x$ onto $\mathcal Y$:
\begin{align}
    \bm g_t: \bm x \in \mathbb R^p \mapsto \mathcal Y,\nonumber
\end{align}
where 
\begin{align}
\mathcal Y=\left\{
\begin{array}{cccc}
(1,-\frac{1}{K-1},-\frac{1}{K-1},\ldots,-\frac{1}{K-1}) \\
(-\frac{1}{K-1},1,-\frac{1}{K-1},\ldots,-\frac{1}{K-1}) \\
\vdots   \\
(-\frac{1}{K-1},-\frac{1}{K-1},-\frac{1}{K-1},\ldots,1) \\
\end{array}
\right\} \nonumber
\end{align}
is the collection of all such label vectors in \eqref{eq:y_space}.
% Accordingly, we define a $K$-tuple of separating functions $\bm f(\bm x) = \{f_1(\bm x),f_2(\bm x),\ldots,f_K(\bm x)\}^\T$ with $f_j(\bm x) = 1$ and other entries equaling $-1/(K-1)$ if the predictive class is $j$. %The function space $\mathcal F$ is the same as in \eqref{eq:labelspace}.
% In the prediction stage, $\A$ predicts the label of a feature vector $\bm x$ by $\argmax_{t'=1,\ldots,K} \sum_{t'=1}^t\bm g_{t'}(\mathbf x)$.

What remains in~\eqref{opti} is to determine the weak classifier $\gb_t$ and what its weight $\alpha_t$ should be. 
Under the forward stage-wise optimization framework, at iteration $t$,
$\gb_t$ and $\alpha_t$ are optimized individually, and passes the corresponding \textbf{ignorance score} $\wb_t = [w_{t,1},w_{t,2},\ldots,w_{t,n}]^\T$ to the next round of iteration. \cite{hastie2009multi} showed that optimizing (\ref{opti}) is equivalent to solving the following problem at each round $t$,
\begin{align*}
(\alpha_t,\bm g_t)&=	\argmin_{ \bm g \in \F_0,\ \alpha_t\in \mathbb{R}} \sum_{i=1}^n  w_{t,i}\exp\biggl(-\frac 1 K \alpha_t \mathbf y_i^\T \bm g(\mathbf x_i)\biggr),\\
 w_{t,i} &= \exp\biggl(-K^{-1}\cdot \mathbf y_i^\T  \sum_{j=1}^{t-1}\alpha_j\bm g_j(\mathbf x_i) \biggr),
%\label{eq:singleada_opt}
\end{align*}
%To solve this problem, for notational convenience, we define the function 
%\begin{align}
%	M_t: \mathbf x \mapsto k, \label{eq_Mt}
%\end{align} 
%where $k$ is such that $(\bm g_t(\mathbf x))_{k} = 1$; alternatively, $M_t = \argmax_{k=1,\ldots,K} (\bm g_t)_{k}$. 
and the solution of $\bm g_t$ is 
%of \eqref{eq:singleada_opt} can be solved as
$$ %$\begin{align}
 \bm g_t = \argmin_{\bm g \in \F_0} \sum_{i=1}^n  w_{t,i}\mathbb I\{\gb(\mathbf x_i) \neq \yb_i\} .
$$ %\end{align}
% The reward from this output is then defined in Algorithm~\ref{algo_supervise}.
%CHECK!!
% We define $\wb_t$ as the \textbf{ignorance score} at iteration $t$.
Note that in the above single-agent formulation, there is only one dataset $\bm X$.

\subsubsection{ASCII for the multi-agent case}
In our multi-agent assisted learning scenario, there are multiple datasets $\bm X^{(m)} = [\xb_i^{(1)},\xb_i^{(2)},\ldots,\xb_i^{(n)}]$ (recall Subsection~\ref{subsec_notation}).
Ideally, an agent $\A $ would collate the distributed data into one (according to sample IDs). 
However, in our context, it is not realistic to obtain $\fb$ since data are privately held by each learner.
Our general idea is to define an objective function that apparently involves all the data, but it actually only requires each learner to model on its data, and interchange some summary statistics. 

%As we introduced in Section~\ref{sec_intro}, the main idea of assisted classification is to exchange certain ignorance scores as side information among agents, so that an agent $\A$ can iteratively improve its predictive performance. %Though our goal of assisted learning is to securely integrate information from different data sources. %, while  the goal of boosting methods is to aggregate weak learning models into a robust ensemble model, they are conceptually related since  both will iteratively improve an agent's predictive performance.
%In light of this, we will borrow 

Our goal (for the agent $\A$) is to minimize the following optimization problem 
\begin{align}
\min_{\fb \in \F} \sum_{i=1}^n \ell\bigl( \yb_i, \fb(\xb_i^{(1)}, \ldots,\xb_i^{(M)})\bigr).
\label{opti2}
\end{align}
To develop an operational algorithm, we still use the exponential loss $ \ell: (\mathbf y,\bm f) \mapsto e^{-\frac 1 K \mathbf y^\T \bm f}$,
and the following model class
\begin{align}
\F = \biggl\{\bm f(\mathbf x_i^{(1)}, \ldots,\mathbf x_i^{(M)}) = \sum_{t=1}^T   \sum_{m=1}^{M}\alpha^{(m)}_t\bm g^{(m)}_t(\xb_i^{(m)}) \biggr\},
	\nonumber %\label{additive}
\end{align}
where $\alpha^{(m)}_t\in \R , \, \bm g^{(m)}_t\in \F_0^{(m)}$.
However, optimization of \eqref{opti2} is intractable because of multiple local models $\gb_t^{(m)},\,t=1,2,\ldots,T,\,m=1,2,\ldots,M$. 
To do this, 
we let $\wb_{t}^{(m)} = [w_{t,1}^{(m)},w_{t,2}^{(m)},\ldots,w_{t,n}^{(m)}]^\T$
% = \exp\{\sum_{t'=1}^{t-1}\sum_{m'=1}^M \alpha_{t'}^{(m')}\gb_{t'}^{(m')}(\xb_i^{(m')})+\sum_{m'=1}^{m-1}\alpha_{t}^{(m')}\gb_{t}^{(m')}(\xb_i^{(m')})\}$
be the \textbf{ignorance score} of learner $m$ at iteration $t$, and solve the optimization problem
\begin{align}
	\min_{\gb_t^{(m)} \in \F_0^{(m)},\,\alpha_t^{(m)}\in \mathbb R} \sum_{i=1}^n w_{t,i}^{(m)}e^{-\frac 1 K \alpha_t^{(m)}\yb_i^\T \gb_t^{(m)}(\xb_i^{(m)})}\nonumber
\end{align}
for learner $m$, $m=1,2,\ldots,M$.
We will show that the above model class leads to an iterative interchange protocol that does not depend on the collated data (in Section~\ref{sec_AL2}).
Consequently, the objective in (\ref{opti2}) can be \textit{virtually implemented} in an iterative manner so that agents only need to transmit ignorance scores without data centralization. %sample weights
%Though it seems that all data are still used in one objective function, the optimization won't be applied to collated data.
%We will show in Section~\ref{sec_AL2} that \eqref{opti2} can be \textit{virtually implemented} in an iterative manner so that one agent only needs to send a vector of ignorance scores %sample weights to the next agent in a chain.

%As we introduced in Section~\ref{sec_intro}, the main idea of assisted classification is to exchange certain ignorance scores as side information among agents, so that an agent $\A$ can iteratively improve its predictive performance. %Though our goal of assisted learning is to securely integrate information from different data sources. %, while  the goal of boosting methods is to aggregate weak learning models into a strong ensemble model, they are conceptually related since  both will iteratively improve an agent's predictive performance.
%In light of this, we will borrow 
For the convenience of reading, we show the notation table in Table~\ref{tab:notation}. Note that the table is for multi-agent cases. For two-agent cases, we denote two agents as $\A$ and $\B$, the feature matrices as $\Xb^{(\A)}$ and $\Xb^{(\B)}$, and similarly other notation.

\begin{table}[tb]
    \centering
    \caption{Summary of the notation in the multi-agent scenario. For the two-agent scenario, the sub/sup-scripts (1) and (2) are replaced with (a) and (b), respectively. }
     \label{tab:notation}
	\begin{tabular}{ p{2.5cm}  p{5.5cm} }
    \hline
    \textbf{Notation} &  \textbf{Meaning} \\ \hline
    $K$ & number of classes \\ \hline
    $M$ & number of agents \\ \hline
    $\mathbb I\{\cdot\}$ & Indicator function  (0 or 1) \\ \hline
    $p_m$ & number of features in agent $m$ \\ \hline
    $n$   & number of labels \\ \hline
    $\Xb^{(m)}\in\mathbb R^{n\times p_m}$ & data matrix of agent $m$\\ \hline
    $\yb \in \mathbb R^{n\times K}$ &label matrix %with each row equaling a length $K$-vector
    \\ \hline
    $\gb_t^{(m)}$ & model for agent $m$ and iteration $t$ \\ \hline
    $\F_0^{(m)}$ & model class for $\gb_t^{(m)}$\\ \hline
    $\fb_T$   & ensemble model at round $T$ \\ \hline
    $\F$ & model class for $\fb$ \\ \hline
    $\alpha_t^{(m)}$ & model weight for  $\gb_t^{(m)}$\\ \hline
    $\wb_t^{(m)}\in \mathbb R^n $ & ignorance score for agent $m$ at iteration $t$\\ \hline
      \end{tabular}
\end{table}

%%%%%%%%%%%%%        NEW     SECTION        %%%%%%%%
%\vspace{-0.2cm}
\section{Assisted Classification in Two-Agent Scenarios} \label{sec_AL2}
%\vspace{-0.2cm}

In this section, we consider a  scenario involving two agents, an agent $\A$ that needs side information, and an agent $\B$ that provides assistance. 
Following the notation in Section~\ref{subsec_notation}, we suppose that $\A$ observes $\mathbf x_i^{(\A)} \in \X^{(\A)} $, B observes $\mathbf x_i^{(\B)} \in \X^{(\B)}$, both observe the label $\bm y_i \in \Y$, for $i=1,2,\ldots,n$.
%We also summarize some other frequently used notation in Table~\ref{tab:notation}. 

%\vspace{-0.2cm}
\subsection{Proposed Algorithm} 
%\vspace{-0.2cm}

We first describe the algorithmic procedure for two-agent assisted classification in Algorithm~\ref{algo_ada}.
It is a training procedure for $\A$ and $\B$ to build local models by interchanging ignorance score $\bm w_{t}^{(\B)}$ and $\bm w_{t+1}^{(\A)}$ at each round $t$. At the prediction stage, $\A$ aggregates the prediction result from itself and from $\B$ to produce a final result. 

Algorithm~\ref{algo_ada} describes how $\B$ assists $\A$ iteratively by exchanging ignorance score with $\A$ in each iteration.
We first briefly explain the idea of each step of Algorithm~\ref{algo_ada} below. 
In each iteration, given the current ignorance score, $\A$ first learns a local model, then the corresponding model weight is derived to minimize  $\A$'s in-sample prediction loss. $\A$ calculates the new ignorance score and passes it to $\B$. With the new ignorance score, $\B$ focuses more on the samples that cannot be well-modeled by $\A$, and correspondingly update the local model and the model weight. $\B$ then updates the new ignorance score and passes it to $\A$ at the next round of iteration.

A subroutine of Algorithm~\ref{algo_ada} named Weighted Supervised Training (WST) is concluded in Algorithm~\ref{algo_supervise}. An agent builds a local model by minimizing the weighted in-sample training loss with a specified model class. 

% \jie{introduce using plain language the Algorithm~\ref{algo_ada}, and Algorithm~\ref{algo_supervise} , which is a subroutine of Algorithm~\ref{algo_ada}.}

\begin{algorithm*}[h]
%\vspace{-0.0 cm}
\small
%ASsisted Classification with Ignorance Interchange (ASCII)
\caption{Two-ASCII: Two-Agent Assisted Classification}% (for two agents $\A$ and $\B$)} %Supervised Learning}
\label{algo_ada}
\begin{algorithmic}[1]
\INPUT Labels $\mathbf Y = [\bm y_1,\ldots,\bm y_n]^\T$, data matrix $\mathbf X^{(\A)}$,  loss function $\ell^{(\A)}$ and model class $\F^{(\A)}_0$ (locally/privately held by $\A$), and the counterpart $\mathbf X^{(\B)}$, $\ell^{(\B)}$, $\F^{(\B)}_0$  (locally/privately held by $\B$), a stop criterion (elaborated in Subsection~\ref{subsec_discussion}).
\OUTPUT Learned $\bm g_t^{(\B)} \in \F^{(\B)}_0, \alpha^{(\A)}_t \in \mathbb{R}$ (held by $\A$), $\bm g_t^{(\B)} \in \F^{(\B)}_0, \alpha^{(\A)}_t \in \mathbb{R}$ (held by $\B$), for $t=1,\ldots, T$ where $T$ is the number of rounds.
%\STATE Let $y = [y_1,\ldots,y_n]$, $y = [y_1,\ldots,y_n]$
\STATE {Initialize $\bm w^{(\A)}_1=[w^{(\A)}_{0,1}, \ldots, w^{(\A)}_{0,n}]^\T=[1,\ldots,1]^\T \in \R^n $}

\FOR {$t = 1,2,\ldots$}

\STATE $\A$ learns a local model $\bm g_t^{(\A)}, \mathbf r_t^{(\A)}= \text{WST}(\mathbf y, \mathbf X^{(\A)},\mathbf w_t^{(\A)},\ell^{(\A)},\F^{(\A)}_0)$, where $\text{WST}$ denotes Algorithm~\ref{algo_supervise}.
\STATE $\A$ calculates $\bar r_t^{(\A)} = (\sum_{i=1}^n r_{(t)}^{(\A)})/n$.
\STATE $\A$ calculates the model weight $\alpha^{(\A)}_t$ from \eqrefblue{derive:alpha_a}.
\STATE $\A$ sends weights $\mathbf w^{(\B)}_t = [w^{(\B)}_{t,1}, w^{(\B)}_{t,2},\ldots, w^{(\B)}_{t,n}]^\T$ and $\alpha^{(\A)}_t$ to $\B$, where $w^{(\B)}_{t,i}$ is calculates in~\eqrefblue{derive:weight_b}.
\STATE $\B$ learns a local model $\bm g_t^{(\B)}, \mathbf r_t^{(\B)}=\text{WST}(\mathbf y, \mathbf X^{(\B)},\mathbf w_t^{(\B)},\ell^{(\B)},\F^{(\B)}_0)$.
%\STATE $\B$ calculates $\bar r_t^{(\B)} = (\sum_{i=1}^n r_{(t)}^{(\B)})/n$, break once  $\bar r_t^{(\B)} < 1/K$.
%\STATE Let $n_{a,b} = \sum_{i=1}^n w_{t,i}^{(\B)}r_{t,i}^{(\A)}r_{t,i}^{(\B)},\, n_{\bar a,b} = \sum_{i=1}^n w_{t,i}^{(\B)}(1-r_{t,i}^{(\A)})r_{t,i}^{(\B)},\, n_{a,\bar b} = \sum_{i=1}^n w_{t,i}^{(\B)}r_{t,i}^{(\A)}(1-r_{t,i}^{(\B)}),\, n_{\bar a,\bar b} = \sum_{i=1}^n w_{t,i}^{(\B)}(1-r_{t,i}^{(\A)})(1-r_{t,i}^{(\B)})$, 
% Let $n_{i,j}$ denote the number of data that $r_t^{(\A)}=i$ and $r_t^{(\B)}=j$, for $i,j\in\{0,1\}$.
\STATE $\B$ calculates the model weight $\alpha^{(\B)}_t$ from \eqrefblue{derive:alpha_b}, break if $\alpha^{(\B)}_t < 0$.
\STATE $\B$ calculates 
$
	w^{(\A)}_{t+1,i}
$
in~\eqrefblue{derive:weight_a}.
\STATE $\B$ sends $\mathbf w^{(\A)}_{t+1}$ and $\alpha^{(\B)}_t$ to $\A$.
%\STATE $\A$ learns the local model $\gb_{t+1}^{(\A)},\mathbf r_{t+1}^{(\A)} = \text{WST}(\mathbf y, \mathbf X^{(\A)},\mathbf w_{t+1}^{(\A)},\ell^{(\A)},\F^{(\A)}_0)$, calculate $\bar r_t = (\sum_{i=1}^n r_{(t+1)}^{(\A)})/n$.
%\STATE \textbf{Break} once $\bar r_t<1/ K.$
\ENDFOR
\STATE In the prediction stage, $\A$ predicts a future data label using $\argmax_{k=1,\ldots,K} (\bm p^{(\A)}_k + \bm p^{(\B)}_k)$, where $\bm p^{(\A)}_k =\sum_{t=1}^T \alpha^{(\A)}_t \bm g_t^{(\A)} (x^{(\A)})$ is evaluated by $\A$, and $\bm p^{(\B)}_k) =\sum_{t=1}^T \alpha^{(\B)}_t \bm g_t^{(\B)} (x^{(\B)})$ is evaluated by $\B$ and sent to $\A$.
\end{algorithmic}
\vspace{-0.0cm}%
\end{algorithm*}
 
 %%%%    Algorithm 1        %%%%%%% 
\begin{algorithm*}[h]
\vspace{-0.0 cm}
\small
\caption{WST: Weighted Supervised Training (Subroutine of Algorithm~\ref{algo_ada})} %Supervised Learning}
\label{algo_supervise}
\begin{algorithmic}[1]
\INPUT Observations of $\Yb=[\yb_1,\yb_2,\ldots,\yb_n],\, \mathbf X = [\mathbf x_1,\mathbf x_2,\ldots,\mathbf x_n]$ and $\mathbf  w = [ w_1, w_2,\ldots, w_n]^\T$,
 loss function $\ell$, supervised model class $\F_0$ (from $\X $ to $\Y$).
\OUTPUT  Supervised function $\hat{f}:   \X \mapsto \Y$, reward vector $r$.
\STATE Solve 
 $%\begin{align}
 	\bm g_n = \argmin_{\bm g\in \F_0} \sum_{i=1}^n w_i \ell(\yb_i,\bm g(\mathbf x_i)). %\nonumber	
 $ %\end{align}
\STATE Calculate reward $\mathbf r= [r_1,r_2,\ldots,r_n]^\T \in \{0,1\}^n$ where $r_i=  \mathbb I\{{\gb_n(\mathbf x_i) = \yb_i}\}$.
\end{algorithmic}
\vspace{-0.0cm}%
\end{algorithm*}
%%%%%% end  of Algo    %%%%%%%%%

\subsection{Technical Details of Algorithm~\ref{algo_ada}}

In the following, we introduce the technical aspects of the algorithms.
First, we introduce Algorithm~\ref{algo_supervise}, which is a subroutine of Algorithm~\ref{algo_ada}. 
% \jie{The following proposition indicates that ...}
The following proposition indicates that if the aforementioned exponential loss is used for the empirical risk minimization, each update of the local model is to minimize the average classification error weighted by the ignorance score.
%For an input label $\yb$, a feature matrix $\Xb$, the model class $\F_0$, and sample weights $\wb$, the output will be shown as:
\begin{proposition}\label{prop1:function_g}
Given label $\Yb=[\yb_1,\yb_2,\ldots,\yb_n]$, feature matrix $\Xb=[\xb_1,\xb_2,\ldots,\xb_n]$, the model class $\F_0$, ignorance score $\wb=[w_1,w_2,\ldots,w_n]^\T$, and $\alpha >0$, the minimization problem
\begin{align}
	\min_{\gb \in \F_0} \sum_{i=1}^n w_i \exp\{-\frac{1}{K} \alpha \yb_i^\T \gb(\xb_i) \}, \nonumber
	\end{align} 
	%the output of WST (Algorithm~\ref{algo_supervise}) 
is equivalent to the following problem
\begin{align}
% \gb_0 = 
\argmin_{\gb \in \F_0} \sum_{i}w_i \mathbb I\{\yb_i \neq \gb(\xb_i)\}.\nonumber  	
\end{align}
\end{proposition}
% \jie{From the proposition, ... the reward ... to simplify later derivations. It describes ...}
From the proposition, the update of the local model $\gb_n$ is only determined by the current ignorance score and the prediction reward from this model.
The reward is a length-$n$ vector that describes how model performs on each sample. To simplify later derivations, we describe  
the reward $\mathbf r = \{r_1,r_2.\ldots,r_n\}^\T$ as
\begin{align}
	r_i = \mathbb I\{\gb_n(\xb_i)=\yb_i\}.\nonumber
\end{align}

Next, we introduce Algorithm~\ref{algo_ada} for the two-agent case.
%We explicitly explain Algorithm~\ref{algo_ada} for the convenience of understanding.
% Suppose we estimate $\fb$ in  (\ref{opti2}) with additive models in the form of  (\ref{additive}).
% Our derivations are based on the following assumptions.
% \jie{
% A1) ...
% A2) ...}
% \begin{itemize}
% 	\item $\alpha_t^{(\A)}$ and  $\alpha_t^{(\B)}$ are positive.
% 	\item $\Xb^{(\A)}\in \mathbb R^{n\times p_{\A}}$ and $\Xb^{(\B)}\in \mathbb R^{n\times p_{\B}}$, $\yb \in \mathbb R^n$.
% \end{itemize}
We adopted a forward stage-wise approach that %consists of two key steps.
  %First, 
  at each round $t$, we fix the already learned parameters and additive components from rounds $1,\ldots, t-1$, and obtain the optimization problem
\begin{align}
		&\min_{\gb_t^{(\A)},\gb_t^{(\B)},\alpha_t^{(\A)},\alpha_t^{(\B)}}\sum_{i=1}^n \ell\biggl(\bm y_i,  \bm g_{0:t-1}(\mathbf x_i^{(\A)},\mathbf x_i^{(\B)})\nonumber \\
		&\quad  + \alpha^{(\A)}_t \bm g^{(\A)}_t(\mathbf x_i^{(\A)}) + \alpha^{(\B)}_t \bm g^{(\B)}_t(\mathbf x_i^{(\B)})\biggr),
		\label{AL1}
\end{align}
where 
\begin{align*}
&\bm g_{0:t-1}(\mathbf x_i^{(\A)},\mathbf x_i^{(\B)})  \\
&= \sum_{j=1}^{t-1}
\biggl(
\alpha^{(\A)}_j \bm g^{(\A)}_j(\mathbf x_i^{(\A)}) + \alpha^{(\B)}_j \bm g^{(\B)}_j(\mathbf x_i^{(\B)}) 
\biggr)
\end{align*}
denotes the already learned model at round $t-1$, and we initialize $\bm g_{0:0}: (\bm x^{(\A)}, \bm x^{(\B)}) \mapsto \bm 0 \in \mathbb R^{K}$. 

Unlike the single-agent case, solving the optimization problem in (\ref{AL1}) is usually intractable because there exist multiple models that need to be simultaneously optimized. 
Under a forward stage-wise additive modeling scheme, for agent $\A$ at iteration $t$, we optimize $\gb_t^{(\A)}$ and $\alpha_t^{(\A)}$ from objective
 \begin{align}
	\min_{\gb_t^{(\A)},\alpha_t^{(\A)}}	\sum_{i=1}^n \ell\biggl(\bm y_i,  \bm g_{0:t-1}(\mathbf x_i^{(\A)},\mathbf x_i^{(\B)}) 
		  + \alpha^{(\A)}_t \bm g^{(\A)}_t(\mathbf x_i^{(\A)} )\biggr) 
		  \label{opt:for_A}
\end{align}
first, we then optimize $\gb_t^{(\B)}$ and $\alpha_t^{(\B)}$ from objective
\begin{align}
		&\min_{\gb_t^{(\B)},\alpha_t^{(\B)}}\sum_{i=1}^n \ell\biggl(\bm y_i,  \bm g_{0:t-1}(\mathbf x_i^{(\A)},\mathbf x_i^{(\B)})\nonumber \\
		&\quad  + \alpha^{(\A)}_t \bm g^{(\A)}_t(\mathbf x_i^{(\A)}) + \alpha^{(\B)}_t \bm g^{(\B)}_t(\mathbf x_i^{(\B)})\biggr)
		\label{opt:for_B}
\end{align}
after $\gb_t^{(\A)}$ and $\alpha_t^{(\A)}$ are determined.
%The algorithm updates the model parameter $\alpha_t^{(\A)}$ first and optimizes $\alpha_t^{(\B)}$ next in each round of optimization. 

Let $w_{t,i}^{(\A)}$ be the ignorance score that is derived at iteration $t-1$,
\eqref{opt:for_A} becomes 
\begin{align}
	\min_{\bm g^{(\A)}_t,\alpha_t^{(\A)}} \sum_{i=1}^n w_{t,i}^{(\A)}  \exp\{-\frac 1 K \yb_i^{\T} \alpha^{(\A)}_t \bm g^{(\A)}_t(\mathbf x_i^{(\A)})
	\}.
	 \label{opt:a} 
\end{align}
and \eqref{opt:for_B} becomes
 \begin{align}
	\min_{\bm g^{(\B)}_t,\alpha_t^{(\B)}} \sum_{i=1}^n & w_{t,i}^{(\B)}\exp\biggl(-\frac 1 K \yb_i^{\T} \{\alpha^{(\A)}_t \bm g^{(\A)}_t(\mathbf x_i^{(\A)}\nonumber\\
	&+\alpha^{(\B)}_t \bm g^{(\B)}_t(\mathbf x_i^{(\B)})\}
	\biggr).
	\label{opt:b}
\end{align}
The ignorance scores in \eqref{opt:a} and \eqref{opt:b} are $\wb_t^{(\A)}$
and $\wb_t^{(\B)}$, respectively. Recall that $\gb_t^{(\A)}$ and $\gb_t^{(\B)}$ are derived from Algorithm~\ref{algo_supervise}. The remaining part is to derive the rule of updating $\alpha_t^{(\A)},\,\alpha_t^{(\B)}$ as well as ignorance scores $\wb_t^{(\B)}$ and $\wb_{t+1}^{(\A)}$. Note that $\wb_{t+1}^{(\A)}$ is the ignorance score $\B$ passes to $\A$ such that $\A$ can initialize the next round of iteration.

We summarize the parameter-updating rule of Algorithm~\ref{algo_ada} in Proposition~\ref{prop2:algo1_rule}.

\begin{proposition}   \label{prop2:algo1_rule}
Given label matrix $\Yb$, covariate matrices $\Xb^{(\A)}$, $\Xb^{(\B)}$, model class $\F_0^{(\A)}$, $\F_0^{(\B)}$ and $\ell^{(\A)}$, $\ell^{(\B)}$ in $\A$ and $\B$, 
under the forward stage-wise additive modeling scheme, the optimal parameters of $\alpha_t^{(\A)}$, $\wb_{t}^{(\B)}$, $\alpha_{t}^{(\B)}$, and $\wb_{t+1}^{(\A)}$ at each iteration $t$ in Algorithm~\ref{algo_ada} are given by the following equations.
\begin{align}
    \alpha_t^{(\A)} & = \bigl\{\log(\bar{r}_t^{(\A)}/ (1-\bar{r}_t^{(\A)}))+\log(K-1)\bigr\} \label{derive:alpha_a}
\end{align}
\begin{align}
w_{t,i}^{(\B)} &= \frac{w_{t,i}^{(\A)} e^{(1-r_{t,i}^{(\A)})\alpha^{(\A)}_t}}{\sum_{i=1}^n w_{t,i}^{(\A)} e^{(1-r_{t,i}^{(\A)})\alpha^{(\A)}_t}} \label{derive:weight_b} \\
\alpha^{(\B)}_t &= \log(K-1)+\log\bigl(e^{\frac{\alpha_t^{(\A)}}{(K-1)^2}}n_{\bar \A, \B}+e^{-\frac{\alpha_t^{(\A)}}{(K-1)}}n_{\A, \B}\bigr) \nonumber\\
& \quad - \log\bigl(e^{\frac{\alpha_t^{(\A)}}{(K-1)^2}}n_{\bar \A,\bar \B}+e^{-\frac{\alpha_t^{(\A)}}{(K-1)}}n_{ \A, \bar \B}.\bigr) \label{derive:alpha_b}\\
w_{t+1,i}^{(\A)} &= \frac{w_{t,i}^{(\B)} e^{(1-r_{t,i}^{(\B)})\alpha^{(\B)}_t}}{\sum_{i=1}^n w_{t,i}^{(\B)} e^{(1-r_{t,i}^{(\B)})\alpha^{(\B)}_t}} \label{derive:weight_a}	
\end{align}
where $\bm g^{(\A)}_t,\, \mathbf r_t^{(\A)}$ and $\bm g^{(\B)}_t,\, \mathbf r_t^{(\B)}$ are Algorithm~\ref{algo_supervise}'s outputs, with inputs $\Yb$, $\mathbf X^{(\A)}$, $\mathbf w_t^{(\A)}$, $\ell^{(\A)}$, $\F^{(\A)}_0$ and $\Yb$, $\mathbf X^{(\B)}$, $\mathbf w_t^{(\B)}$, $\ell^{(\B)}$, $\F^{(\B)}_0$, respectively.
 Moreover, 
\begin{align*}
\bar{r}_t^{(\A)} = \frac{\sum_{i=1}^n w_{t,i}^{(\A)}r_{t,i}^{(\A)}}{\sum_{i=1}^n w_{t,i}^{(\A)}},
\end{align*}
and 
$
n_{\A,\B} = \sum_{i=1}^n w_{t,i}^{(\B)}r_{t,i}^{(\A)}r_{t,i}^{(\B)},\, n_{\bar \A,\B} = \sum_{i=1}^n w_{t,i}^{(\B)}(1-r_{t,i}^{(\A)})r_{t,i}^{(\B)},\, n_{\A,\bar \B} = \sum_{i=1}^n w_{t,i}^{(\B)}r_{t,i}^{(\A)}(1-r_{t,i}^{(\B)}),\, n_{\bar \A,\bar \B} = \sum_{i=1}^n w_{t,i}^{(\B)}(1-r_{t,i}^{(\A)})(1-r_{t,i}^{(\B)})$. 

 \end{proposition}

  %A similar formulation for a single agent  has been used in the literature. However, 
%\eqref{AL1} cannot be optimized because there exist two parameters $\alpha^{(\A)}_t,\,\alpha^{(\B)}_t$. 
The above proposition describes how $\A$ interchanges with $\B$ in each iteration.
Next, we provide a sketch proof and technical discussions on how the problem \eqref{AL1} is solved by interchanging ignorance scores. % under the forward additive scheme. 

The basic idea of Algorithm~\ref{algo_ada} is to first optimize $\alpha^{(\A)}_t$ after getting $\gb_t^{(\A)}$ from Algorithm~\ref{algo_supervise}, then $\A$ passes the ignorance score  to $\B$, $\B$ optimizes $\alpha_t^{(\B)}$ after getting $\gb_t^{(\B)}$ from Algorithm~\ref{algo_supervise}, and finally $\B$ passes the updated ignorance score back to $\A$ for next round of interchange.

Thus, we first consider the optimization for $\A$.
At line 3, we obtain $\bm g^{(\A)}_t$ and $\bm r_t^{(\A)} = [ r_{t,1}^{(\A)},r_{t,2}^{(\A)},\ldots,r_{t,n}^{(\A)}]^\T$ from the output of $\text{WST}(\yb,\Xb^{(\A)},\wb_t^{(\A)},\ell^{(\A)},\F_0^{(\A)})$.
At line 5, the update of $\alpha_t^{(\A)}$ is optimized based on \eqref{opt:a} which results in \eqref{derive:alpha_a}.
At line 6, $\A$ then calculates the ignorance score $\wb_{t,i}^{(\B)}$ 
in \eqref{derive:weight_b}
and passes it to $\B$.

For agent $\B$, similarly, at line 7, we derive $\bm g^{(\B)}_t$ and $\bm r_t^{(\B)} = [ r_{t,1}^{(\B)},r_{t,2}^{(\B)},\ldots,r_{t,n}^{(\B)}]^\T$ from the output of $\text{WST}(\yb,\Xb^{(\B)},\wb_t^{(\B)},\ell^{(\B)},\F_0^{(\B)})$.
%For agent $\B$, the optimization function becomes \eqref{AL1}. 
When $\alpha_t^{(\A)},\,\gb_t^{(\A)},\,\gb_t^{(\B)}$ are known, the optimization of \eqref{AL1} reduces to \eqref{opt:b}
using the ignorance score $\wb_t^{(\B)}$.
The only remaining task is to find such an $\alpha_t^{(\B)}$ to optimize \eqref{opt:b}.
%Let $n_{\A,\B} = \sum_{i=1}^n w_{t,i}^{(\B)}r_{t,i}^{(\A)}r_{t,i}^{(\B)},\, n_{\bar \A,\B} = \sum_{i=1}^n w_{t,i}^{(\B)}(1-r_{t,i}^{(\A)})r_{t,i}^{(\B)},\, n_{\A,\bar \B} = \sum_{i=1}^n w_{t,i}^{(\B)}r_{t,i}^{(\A)}(1-r_{t,i}^{(\B)}),\, n_{\bar \A,\bar \B} = \sum_{i=1}^n w_{t,i}^{(\B)}(1-r_{t,i}^{(\A)})(1-r_{t,i}^{(\B)})$, at line 8, 
$\alpha_t^{(\B)}$ is then updated based on \eqref{opt:b} and concluded in~\eqref{derive:alpha_b}.
At line 9, $\B$ then calculates the ignorance value $\wb_{t+1}^{(\A)}$ in \eqref{derive:weight_a}
 and passes it to $\A$ for next round of iteration.

\begin{figure*}
    \centering
    \includegraphics[width=1\linewidth]{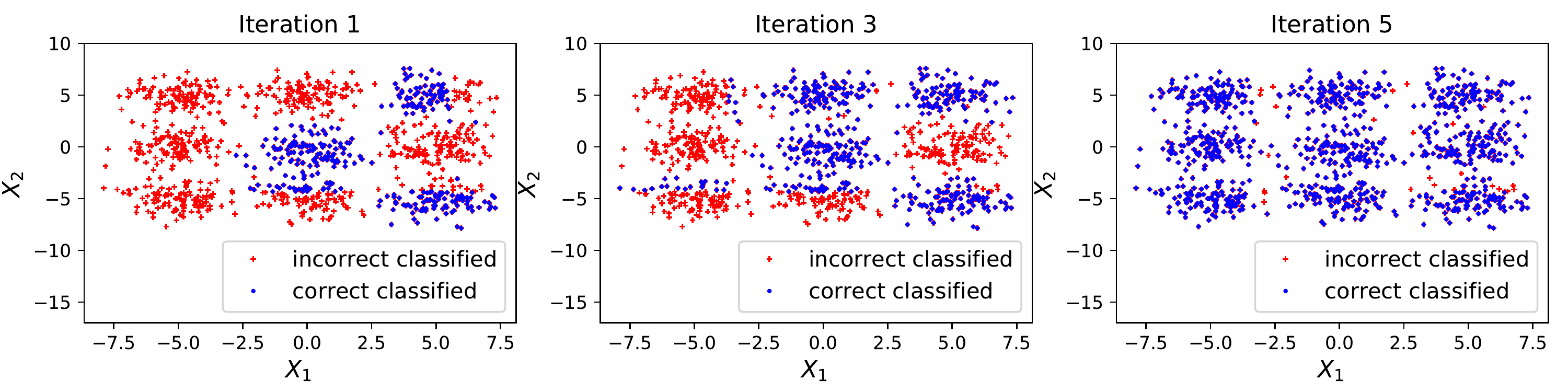}
    %%\vspace{-0.2in}
    \caption{Demonstration on how ASCII works using the Blob data, where each learner holds one feature. The model used by each agent is a decision tree. The blue dots indicate correctly classified points, while red ones are incorrect. In this example, an agent cannot well discriminate all classes on its own, but ASCII helps it to achieve desirable performance.} 
    \label{experiment0}
\end{figure*}
%\vspace{0.1cm}

\subsection{More discussions} \label{subsec_discussion}

In this subsection, we include more discussions on the stop criterion, complexity analysis, and interpretations through some special cases.

\textbf{Stop criteria}.
The stop criterion as an input of Algorithm~\ref{algo_ada} guides $\A $ when to stop the assisted classification with $\B $. 
We suggest two stop criteria for practice use.
With the first stop criterion, the procedure of iterative assistance is repeated $K$ times until 
$ %\begin{align}
    \bar r_t^{(\A)} \leq 1/K. %\label{eq_102}
$ %\end{align}
It can be verified that this condition is equivalent to $\alpha_t^{(\A)}\leq 0$. An insight of the stopping criteria is that when the current local model is worse than random guessing, the algorithm should be terminated.
Our experiments and uploaded codes are based on this criterion. %Various experiments indicate that this criterion

The second stop criterion we suggest is to use the cross-validation technique~\cite{allen1974relationship,geisser1975predictive}.
In particular, $\A $ and $\B $ only use a part of the data (e.g. the first 50\% rows aligned by data IDs) to perform the assisted learning. They preserve the remaining rows to evaluate $\A $'s performance as if in the prediction stage.
The learning process continues until $\A $'s average out-sample predictive error no longer decreases. 
The cross-validation is a general method with theoretical guarantees on the generalization capability ~\cite{DingOverview}.
Nevertheless, the produced predictive performance has been practically and theoretically shown to be sensitive to the training-testing splitting ratio~\cite{DingOverview}.
Also, $\A $ may not be able to do a re-training after the validation procedure due to communication/computation constraints. Due to the above concerns, we used the other criterion in the experiments. 

\textbf{Algorithm complexity}.
%Typical complexity analysis depend on specific type of algorithms.
In each iteration of Algorithm~\ref{algo_ada}, $\A $ and $\B $ each trains a model. This part of complexity depends on the specific models used by $\A $ and $\B $.
Additionally, $\A $ and $\B $ each calculates the ignorance score with computational and storage complexity $O(n)$ (where $n$ is the sample size).
As a result, the overall complexity for $\A $ is $O(nI+c_\A I)$, where $c_\A$ is her own model complexity and $I$ is the number of iterations. Similar complexity applies to $\B $.

\textbf{Interpretation of the derived} $\alpha_t^{(\A)}$ \textbf{and} $\wb_t^{(\A)}$.
The derivation of $\wb_t^{(\A)}$ indicates that the samples not well-modeled by $\A$ will be relatively more focused by $\B$.
Similar arguments apply to $\wb_t^{(\B)}$.
We define $n_{\A} = \sum_{i=1}^n w_{t,i}^{(\A)}r_{t,i}^{(\A)}$ and  $n_{\bar\A} = \sum_{i=1}^n w_{t,i}^{(\A)}(1-r_{t,i}^{(\A)})$.
 At iteration $t$, 
the value of $\alpha_t^{(\A)}$ 
can be represented as 
\begin{align}
	\alpha_t^{(\A)}\approx \log(K-1)+\log \biggl(\frac{n_{\A}}{n_{\bar\A}}\biggr) \nonumber,
\end{align}
which implies that less important samples are mis-classified, a larger model weight will be given. If all the samples are correctly classified by the current model, $\alpha_t^{(\A)}$ becomes infinity. 
The value of $\alpha_t^{(\B)}$ is related to both $\alpha_t^{(\A)}$ and $\B$'s own predictive performance. In particular, when $\alpha_t^{(\A)}$ is close to zero, $\alpha_t^{(\B)}$ will be close to 
\begin{align}
	\alpha_t^{(\B)}&\approx \log(K-1)+\log\biggl(\frac{n_{\bar\A,B}+n_{\A,\B}}{n_{\bar\A,\bar B}+n_{\A,\bar\B}}\biggr)\nonumber\\
	&=\log(K-1)+\log\biggl(\frac{n_{\B}}{n_{\bar\B}}\biggr)\nonumber.
\end{align} 
 
\textbf{Objective privacy}.
While in our algorithm, each agent needs to access the original task label, it does not necessarily leak the task initiator's private learning objective.
More specifically, to receive assistance from others, $\A$ only needs to send the numerical task label, and the semantic information of $\A$'s underlying task is not shared. In this case, ASCII can improve $\A$'s predictive performance while maintaining a private objective.

\textbf{A toy example for interpretations}.
We illustrate how our algorithm works on a nine-class `Blobs' data, and  
snapshot the fitted results in Figure~\ref{experiment0} to illustrate how the algorithm alternates between two agents $\A,\B$ in order to perform $\A$'s learning performance.

%In Sec.. we will provide experiments that show ...

%%%%%%%%%%%%%%%%%%%%%%%%%%%%%%%%%%
%\vspace{-0.2cm}
\section{Extension to Multi-Agent Scenarios}
\label{sec_ALm}
%\vspace{-0.3cm}

\begin{figure*}
\centering
	\includegraphics[width=0.8\linewidth]{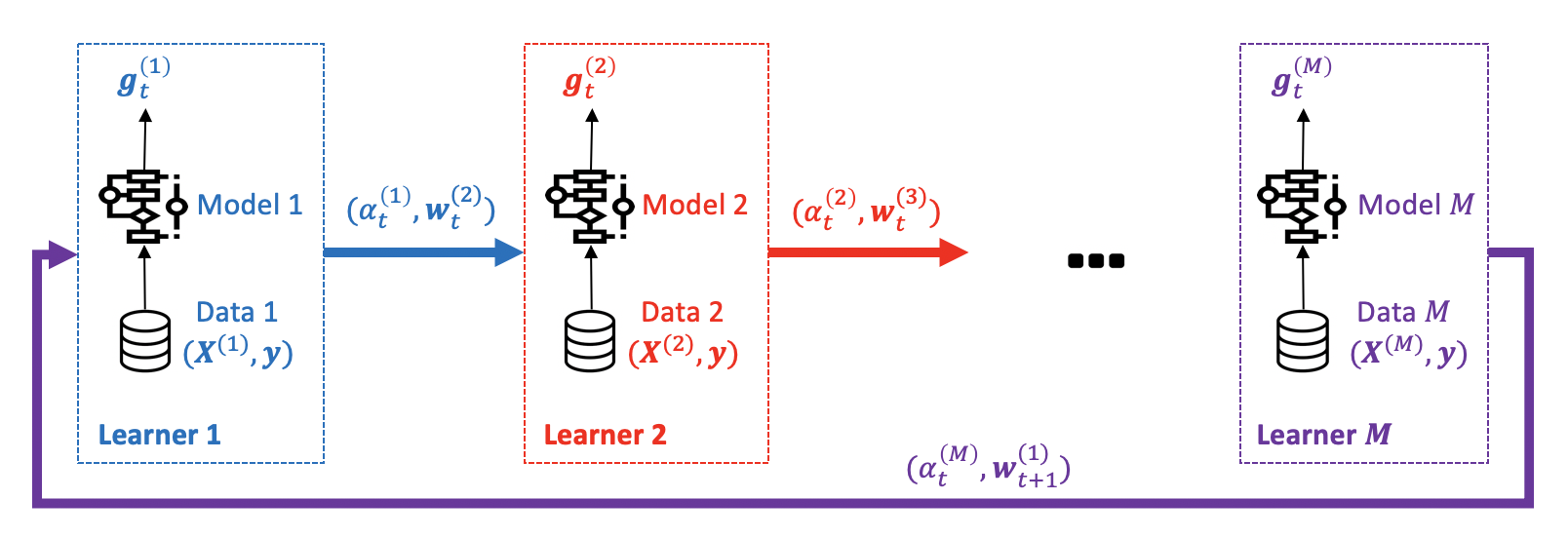}
	\vspace{-0.3cm}
	\caption{Demonstration of the algorithmic update in the presence of $M$ agents.}
	\label{fig:algorithmplot}
\end{figure*}

We now briefly explain the extension of Algorithm~\ref{algo_ada} to  multi-agent scenarios where the objective is given in  \eqref{opti2}.
In addition to the notation introduced in Section~\ref{subsec_notation},
we suppose that the ignorance score is exchanged in a chain of agents, say $1 \to 2 \ldots  \to M $, and the last agent will transmit the score to the first agent, who will then initialize a new round of interactions (demonstrated in Figure~\ref{fig:algorithmplot}). 
We will perform empirical studies to compare the performance from a deterministic chain with a random sequence of agents (with replacement).
An adaptive selection of the order of interchange is left as future work.
% The way that the $i+1$th agent assisted by $i$th leaner is from the transmission of updated data-driven weight $\mathbf w_{t}^{(i+1)}$ and model-driven weight $\alpha_{t}^{(i)},\,k=1,2,\ldots,M-1.$ The updated model weight and data driven weight, which is denoted as $\alpha_{t}^{(M)}$ and $\mathbf w_{t+1}^{(1)}$, will be transmitted from $\mathcal L_M$ to $\mathcal L_1$. The algorithm then kicks off next round of interaction beginning at $\mathcal L_1$.

%In particular, consider the additive model in \eqref{additive}, 
Suppose $M$ agents access the label matrix $\Yb = [\yb_1,\yb_2,\ldots,\yb_n]$, and each agent $m$
possesses a feature matrix $\Xb^{(m)} = [\xb_1^{(m)},\ldots,\xb_n^{(m)}]^\T$, a model class $\F_0^{(m)}$ and a loss function $\ell^{(m)}$ for $m=1,2,\ldots,M$.
At iteration $t$, for agent $m$, under the forward stage-wise additive modeling scheme, denote the already-learned virtually-joint model from previous $t-1$ iterations as  $$\bm g_{0:t-1}(\bm x_i^{(1)},\bm
x_i^{(2)},\ldots,\bm x_i^{(M)}) =  \sum_{\tau=1}^{t-1}\sum_{j=1}^{M}\alpha^{(j)}_\tau \bm g^{(j)}_\tau(\mathbf x_i^{(j)}),$$ where we initialize $\bm g_{0:0}: (\bm x^{(1)},\ldots, \bm x^{(M)})\mapsto \bm 0$. Similar as \eqref{opti2}, the optimization function can be expressed as
 \begin{align}
	\min_{\alpha_t^{(m)}}	\sum_{i=1}^n \ell\biggl(\bm y_i,  \bm g_{0:t-1}(\mathbf x_i^{(1)},\ldots,\mathbf x_i^{(M)}) 
		  + \sum_{j=1}^{m}\alpha^{(j)}_t \bm g^{(j)}_t(\mathbf x_i^{(j)} )\biggr). \nonumber	
\end{align}
Under the forward stage-wise additive modeling scheme, at iteration $t$, agent $m$ receives the ignorance score $ \wb_t^{(m)}$ from agent $m-1$ (or from agent $M$ at iteration $t-1$ if $m=1$), and minimizes the exponential in-sample prediction loss that considers the additive models until $\gb_t^{(m)}$. The objective function for the agent $m$ is
\begin{align}
  	&\min_{\alpha^{(m)}_t\in \R , \, \bm g^{(m)}_t \in \F^{(m)}_0}  \sum_{i=1}^n 
  	w_{t,i}^{(m)} e^{-\frac{1}{K}\bm y_i^\T \bigl\{ \sum_{j=1}^m \alpha^{(j)}_t \bm g_t^{(j)}(\mathbf x^{(j)}_i) \bigr\}}	,   	
  	\nonumber %\label{opti:multiple}
  \end{align}
where $ \wb_t^{(m)}=[w_{t,1}^{(m)},w_{t,2}^{(m)},\ldots,w_{t,n}^{(m)}]^\T $ ,
and $\alpha^{(j)}_t,\,j=1,2,\ldots,m-1$ that have been transferred by $m-1$ preceding agents are known parameters. \eqref{opt:b} is the specific case when $M=2$.
Similarly, $\gb_t^{(m)}$ is learned from the objective function
\begin{align}
\min_{\gb_t^{(m)}\in\F_0^{(m)}}\sum_{i=1}^n w_{t,i}^{(m)}\exp\{-\frac 1 K \alpha_t^{(m)}\yb_i^\T \gb_t^{(m)}(\xb_i^{(m)})\},\nonumber
\end{align}
and can be solved according to Proposition~\ref{prop1:function_g}, which is the output of $\text{WST}(\mathbf Y, \mathbf X^{(m)},\mathbf w_t^{(m)},\ell^{(m)},\F^{(m)}_0)$.

Consider the agent $m$ at iteration  $t$, the side information from previous iterations are already reflected in $ \wb_t^{(m)}$,
The above objective function can be rewritten as
\begin{align}
    \min_{\alpha_t^{(m)}, g_t^{(m)}}
    \biggl\{
    &\sum_{I_{t}^{(m)}}
  	w_{t,i}^{(m)} e^{-\frac{1}{K}\bm y_i^\T \{ \sum_{j=1}^{m-1} \alpha^{(j)}_t \bm g_t^{(j)}(\mathbf x^{(j)}_i) \}}e^{-\frac{1}{K-1}\alpha_t^{(m)}} \nonumber \\
  	&\hspace{-1cm}+\sum_{I_{\bar t}^{(m)}}
  	w_{t,i}^{(m)} e^{-\frac{1}{K}\bm y_i^\T \{ \sum_{j=1}^{m-1} \alpha^{(j)}_t \bm g_t^{(j)}(\mathbf x^{(j)}_i) \}}e^{\frac{1}{(K-1)^2}\alpha_t^{(m)}} \nonumber
  	\biggr\},
\end{align}
where $\gb_t^{(m)}$ is implicitly defined in $I_{t}^{(m)}$ and $I_{\bar t}^{(m)}$, where $I_{t}^{(m)} = \bigl\{i:\gb_t^{(m)}(\bm x_i^{(m)}) = \yb_i \bigr\}$ and $I_{\bar t}^{(m)} = \bigl\{i:\gb_t^{(m)}(\bm x_i^{(m)}) \neq \yb_i \bigr\}$.

It can be verified that the above function is convex in $\alpha_t^{(m)}$. Taking the derivative with respect to $\alpha_t^{(m)}$, and letting it be zero, we obtain the optimum at
\begin{equation}
	\begin{aligned}
    \alpha_t^{(m)}& = \biggl\{\log \frac{\sum_{I_t}
  	w_{t,i}^{(m)} e^{-\frac{1}{K}\bm y_i \bigl\{ \sum_{j=1}^{m-1} \alpha^{(j)}_t \bm g_t^{(j)}(\mathbf x^{(j)}_i) \bigr\}}}{\sum_{I_{\bar t}}
  	w_{t,i}^{(m)} e^{-\frac{1}{K}\bm y_i \bigl\{ \sum_{j=1}^{m-1} \alpha^{(j)}_t \bm g_t^{(j)}(\mathbf x^{(j)}_i) \bigr\}}} \\
  	&\quad + \log(K-1)\biggr\}\times\frac{K}{(K-1)^2}. 
\end{aligned}
 	\label{eq_101}
\end{equation}
Since $K/(K-1)^2$ is a constant for all  $\alpha_t^{(m)},\,t=1,2,\ldots,\,m=1,2,\ldots,M$, it can be removed  from (\ref{eq_101}) without changing the algorithmic output.
From \eqref{eq_101}, $\alpha_t^{(m)}$ will be larger if more important samples (with larger $w_{t,i}^{(m)}$) are correctly classified by $\gb_t^{(m)}$.
We then update the ignorance score. Figure~\ref{fig:algorithmplot} describes how each agent interchanges with each other. If $m \neq M$, the ignorance that agent $m$ transmits to $m+1$ is
\begin{align}
    w_{t,i}^{(m+1)} = \frac{w_{t,i}^{(m)}  \times \exp\{\alpha_t^{(m)} (1-r_{t,i}^{(m)})\}}{\sum_{i=1}^n w_{t,i}^{(m)}  \times \exp\{\alpha_t^{(m)} (1-r_{t,i}^{(m)})\}}\nonumber
\end{align}
 for $m >0$. If $M=1$, the new ignorance score will be transmitted to agent $1$ at next round of iteration, which is
 \begin{align}
    w_{t+1,i}^{(1)} = \frac{w_{t,i}^{(M)}  \times \exp\{\alpha_t^{(M)} (1-r_{t,i}^{(M)})\}}{\sum_{i=1}^n w_{t,i}^{(M)}  \times \exp\{\alpha_t^{(M)} (1-r_{t,i}^{(M)})\}}.\nonumber
    % \label{eq:transfer weight}
\end{align}
The interchanges at different iterations are therefore connected between agent $M$ and agent $1$.
% \begin{align}
%     \alpha_t^{(m)} = \log \frac{\sum_{I_b}
%   	w_{t,i}^{(m)} e^{-\frac{1}{K}\bm y_i \bigl\{ \sum_{j=1}^{m-1} \alpha^{(j)}_t \bm g_t^{(j)}(\mathbf x^{(j)}_i) \bigr\}}}{\sum_{I_{\bar b}}
%   	w_{t,i}^{(m)} e^{-\frac{1}{K}\bm y_i \bigl\{ \sum_{j=1}^{m-1} \alpha^{(j)}_t \bm g_t^{(j)}(\mathbf x^{(j)}_i) \bigr\}}} + \log(K-1).
%   	\label{multiple:alpha}
% \end{align}

%%%%%%%%%.  Similar Method.     %%%%%%%%%%%%%%
\section{Other Variants} \label{sec_other}

In this section, we also consider some variants of the ASCII method developed in previous sections, which also aim to solve \eqref{opti2}.
They will be experimentally compared in Section~\ref{sec_exp}.
%  we derive several related methods regarding this problem. 

\textbf{Method 1} (ASCII-Simple). The first method is similar to ASCII when transferring ignorance scores as described in~Equations \eqref{derive:weight_b} and \eqref{derive:weight_a}, and the only difference is that the update of $\alpha_t^{(m)}$ is based on $m$th agent's individual exponential loss (in the line 8 of Algorithm~\ref{algo_ada}). 
Take the two-agent Algorithm~\ref{algo_ada} as an example. The pseudocode for ASCII-Simple is briefly summarized below.
\begin{itemize}
\item $\A$ learns $\gb_t^{(\A)}$ and $\alpha_t^{(\A)}$ based on the optimization function \eqref{opt:a}.
\item $\A$ calculates the ignorance score $\wb_t^{(\B)}$ as in \eqref{derive:weight_b} and passes it to $\B$.
\item $\B$ learns $\gb_t^{(\B)}$ and $\alpha_t^{(\B)}$ on the optimization function $\min_{\alpha_t^{(\B)}} \sum_{i=1}^n w_{t,i}^{(\B)}  \exp\{-\frac 1 K \yb_i^{\T} \alpha^{(\B)}_t \bm g^{(\B)}_t(\mathbf x_i^{(\B)})
	\}$.
\item $\B$ calculates the ignorance score as in \eqref{derive:weight_a} and passes to $\A$ at the next round of iteration.
\end{itemize}
Intuitively, this method may not be as efficient as the ASCII as the side information of model-level performance was not transmitted to accelerate the learning efficiency. It is conceivable that ASCII will perform better when finish same round of iterations, which is observed our experimental results~\ref{sec_exp:compare method}.

%    \item $\B$ calculates the model weight $\alpha^{(\B)}_t= \{\log(\bar{r}_t^{(\B)}/ (1-\bar{r}_t^{(\B)}))+\log(K-1)\}$, where $\bar{r}_t^{(\B)} = \frac{\sum_{i=1}^n w_{t,i}^{(\B)}r_{t,i}^{(\B)}}{\sum_{i=1}^n w_{t,i}^{(\B)}}$.

% \begin{align}
% 	\min_{\bm f_m \in \F^{(m)}} \sum_{i=1}^n \ell\bigl( \bm y_i,  \bm f_m(\mathbf x^{(m)})\bigr)	
% , \quad \textrm{ where }
% 	\ell: (\mathbf y,\bm f_m) \mapsto e^{-\frac 1 K \mathbf y^\T \bm f_m},
% \label{opti:only transfer weight}
% \end{align}
% where
% \begin{align}
% \F^{(m)} = \biggl\{	\bm f_m(\mathbf x^{(m)}) = \sum_{t=1}^T   \alpha^{(m)}_t\bm g^{(m)}_t(\mathbf x^{(m)}) ,\, \alpha^{(m)}_t\in \R , \, \bm g^{(m)}_t\in \F_0^{(m)} \biggr\}.
% 	\label{additive:only transfer weight}
% \end{align}

\textbf{Method 2} (ASCII-Random). The second method is to randomly shuffle the order of interchange at each iteration. 
%In ASCII, the order of agents that updates the weight in each iteration is fixed, which is listed as $1\to 2\to \ldots \to M$. 
In other words, the ASCII-Random uses the order of $\pi_t(1),\ldots, \pi_t(M)$ where $\pi_t$ denotes a random permutation at each iteration $t$. 

\textbf{Method 3} (Ensemble-Adaboost). The third method is that we ignore the interchange between agents, so that each agent learns an Adaboost model and the final prediction uses a majority vote of all the agents.

% \textbf{Remark2} The communication chain 1 to 2 to ... M is neither practical nor efficient in a large-scale system. Also, the information gain is asymmetric. For example, suppose that agent A holds 100 features and agent B holds only 1 feature. Obviously, B will be significantly improved by communicating with A, but not the other way around for A. Therefore, in our future work e.g., to find the most efficient and effective path to communicate, we would only consider the communicating method starting with one fixed agent.

%\textbf{Method 4}{ (Complete Random ASCII)}. The difference between Method 4 and ASCII is that in iteration $t$, we first repeated sample $M$ numbers from $1,2,\ldots, M$. 

%%%%%%%%   Experiment Study    %%%%%%%%
%\vspace{-0.2cm}
\section{Experimental Study}\label{sec_exp}
%\vspace{-0.2cm}
We provide numerical demonstrations of  the proposed method. For the synthetic data, we replicated 20 times for each method. In each replication, we trained on a dataset with size $10^3$, and then tested on a dataset with size $10^5$. We chose a testing size much larger than the training size to produce a fair comparison of out-sample predictive performance~\cite{Ding2018model}. For the real data, we trained on 70\% and tested on 30\% of data, re-sampled with replacement 20 times to average the performance and to evaluate standard errors. The `oracle
score' is the testing
error obtained by the model that is trained
on the pulled data. 

%\vspace{-0.2cm}
\subsection{ASCII for improving accuracy to near-oracle}
%\vspace{-0.2cm}
We test on synthetic and real data to verify that the proposed method can significantly improve the classification performance for a single agent.
The real data considered are listed below.

\noindent$\bullet$\textbf{MIMIC3 Data}.
Medical Information Mart for Intensive Care III~\cite{johnson2016mimic} (MIMIC3) is a comprehensive clinical database that contains  de-identified information for 38,597 distinct adult patients admitted to critical care units between 2001 and 2012 at a Medical Center.
 We focus on the task of predicting if a patient would have an extended Length of Stay ($ > $ 7 days) based on the first 24 hours information. Following the processing procedures in \cite{harutyunyan2017multitask, purushotham2018benchmarking}, we select 16 medical features and 15000 patients. We partition the data into two agents according to the original data sources, with one holding three features and the other holding 12 features.

\noindent $\bullet$ \textbf{QSAR biodegradation Data}. The quantitative structure-activity relationship (QSAR) biodegradation dataset was built in the Milano Chemometrics and QSAR Research Group \cite{mansouri2013quantitative}. The whole dataset has 41 attributes and 2-class labels, with 1055 data in total. We partition the data vertically into two parts for two agents who hold 20 and 21 features.

\noindent $\bullet$ \textbf{Red Wine Quality Data}. The red wine classification dataset has 1600 data, 11 attributes, and 6-class labels \cite{cortez2009modeling}. We partition the data into two agents, the first holding six features, and the second one holding five features.

\noindent $\bullet$ \textbf{Blob Data}  (Synthetic).  We generate isotropic Gaussian blobs for clustering. $\mathbf X \in \mathbb R^{1000\times 8}$ is the feature matrix and $\mathbf c \in \mathbb R^{1000}$ is the 10-class label.
Suppose that there exist four agents, and each holds two non-overlapping columns in $\mathbf X$.
For simplicity, we let each agent use a random forest model with the same number of trees and depth. 
% that A and By hold $X_1$ and $X_2$, respectively. The snapshots of the classification results at three intermediate rounds are in Fig~\ref{fig_1b}.

\begin{figure}[tb]
     \centering
     \includegraphics[width=1\linewidth]{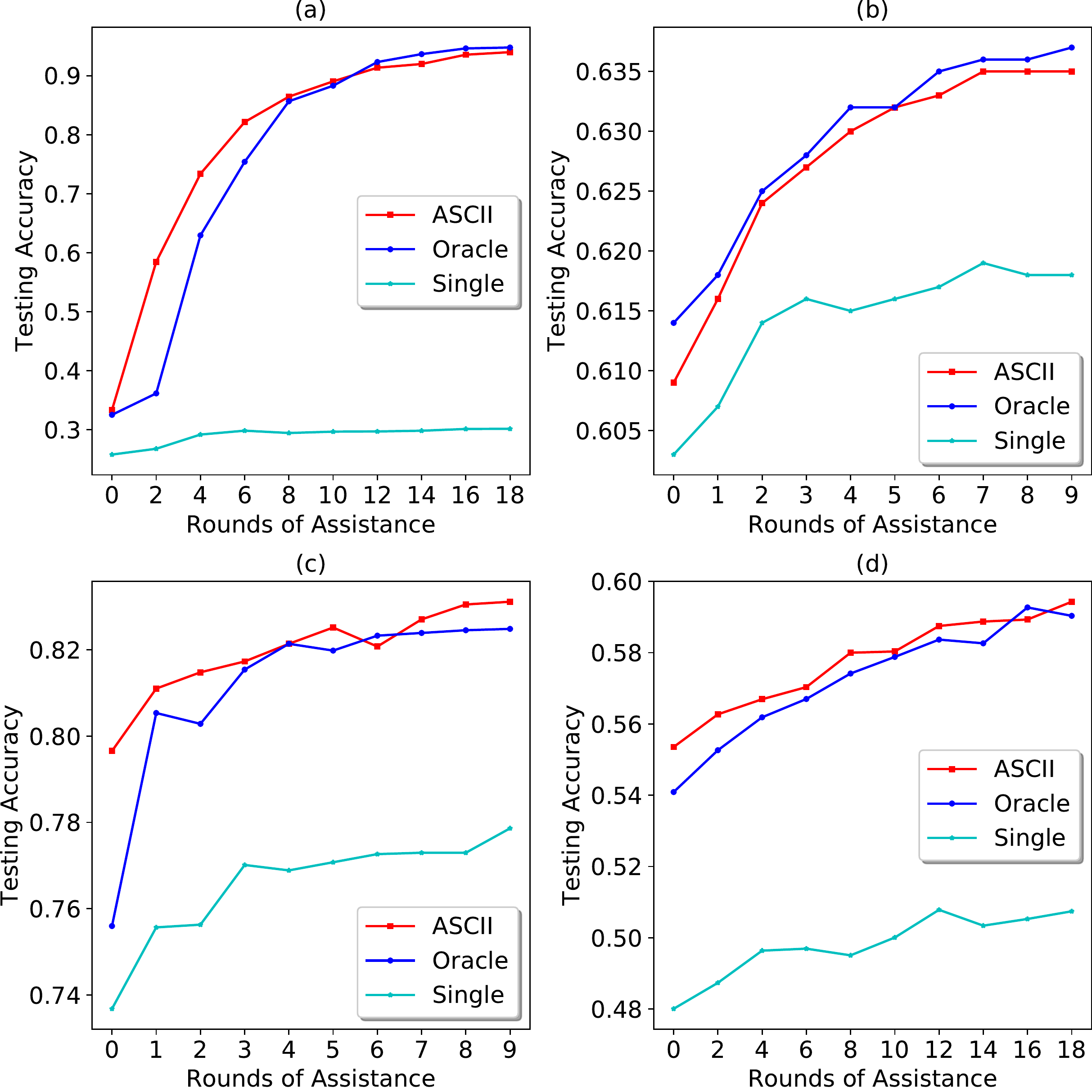}
     \vspace{-0.0in}
     \caption{Out-sample predictive accuracy of the proposed method (`ASCII'), the oracle method unrealistically using the pulled data (`Oracle'), and the non-assisted method using only agent $\A$'s data (`Single'), against the number of rounds  for the datasets of (a) Blob,  (b) MIMIC, (c) QSAR, and (d) Wine, as described in the text. The model used by each agent (a) is random forest, and decision tree in (b)(c)(d). Standard errors over 20 independent replications are within 0.04.}
     \label{experiment1}
     \vspace{-0.0in}
  \end{figure}

We suppose that each agent uses a decision tree classifier for the above data except for the blob data.
The results summarized in Figure~\ref{experiment1} indicate that our proposed method performs significantly better than that of a single agent and often nearly oracle within a few rounds of assistance. 
% The three real data we used are described below.

%%%%%%%%%%%%%%%%%%%%%%
% \vspace{-0.1cm}
% \subsection{Simulated data: 15-class classification of Blob dataset}
% \vspace{-0.1cm}

%     \begin{figure}[tb]
%   \centering
%     \includegraphics[width=1\linewidth]{figures/fig_1b}
%     \vspace{-0.0in}
%     \caption{Illustration of the Moon dataset with two classes (in black dot and blue square), along with the mis-classified points (in red ross) in each of three rounds. }
%     \label{fig_1b}
%     \vspace{-0.0in}
%   \end{figure}
%   
   
%   \begin{figure}[tb]
%   \centering
%      \includegraphics[width=0.7\linewidth]{figures/fig1}
%      \vspace{-0.0in}
%      \caption{Illustration of the testing accuracy (in solid lines) of our method (in dots), the oracle method using all the available data at once (in squares), and the method using only agent $\mathcal L_1$ (in circles).}
%      \label{fig_1a}
%      \vspace{-0.0in}
%   \end{figure}
%  \begin{figure}[!htb]
% \minipage{0.32\textwidth}
%   \includegraphics[width=\linewidth]{figures/mic_claf.pdf}
%   \caption{MIMIC}\label{fig:awesome_image1}
% \endminipage\hfill
% \minipage{0.32\textwidth}
%   \includegraphics[width=\linewidth]{figures/biodeg.pdf}
%   \caption{QSAR}\label{fig:awesome_image2}
% \endminipage\hfill
% \minipage{0.32\textwidth}%
%   \includegraphics[width=\linewidth]{figures/wine.pdf}
%   \caption{Wine quality}\label{fig:awesome_image3}
% \endminipage
% \end{figure}

     \begin{figure}[tb]
     \centering
     \includegraphics[width=1\linewidth]{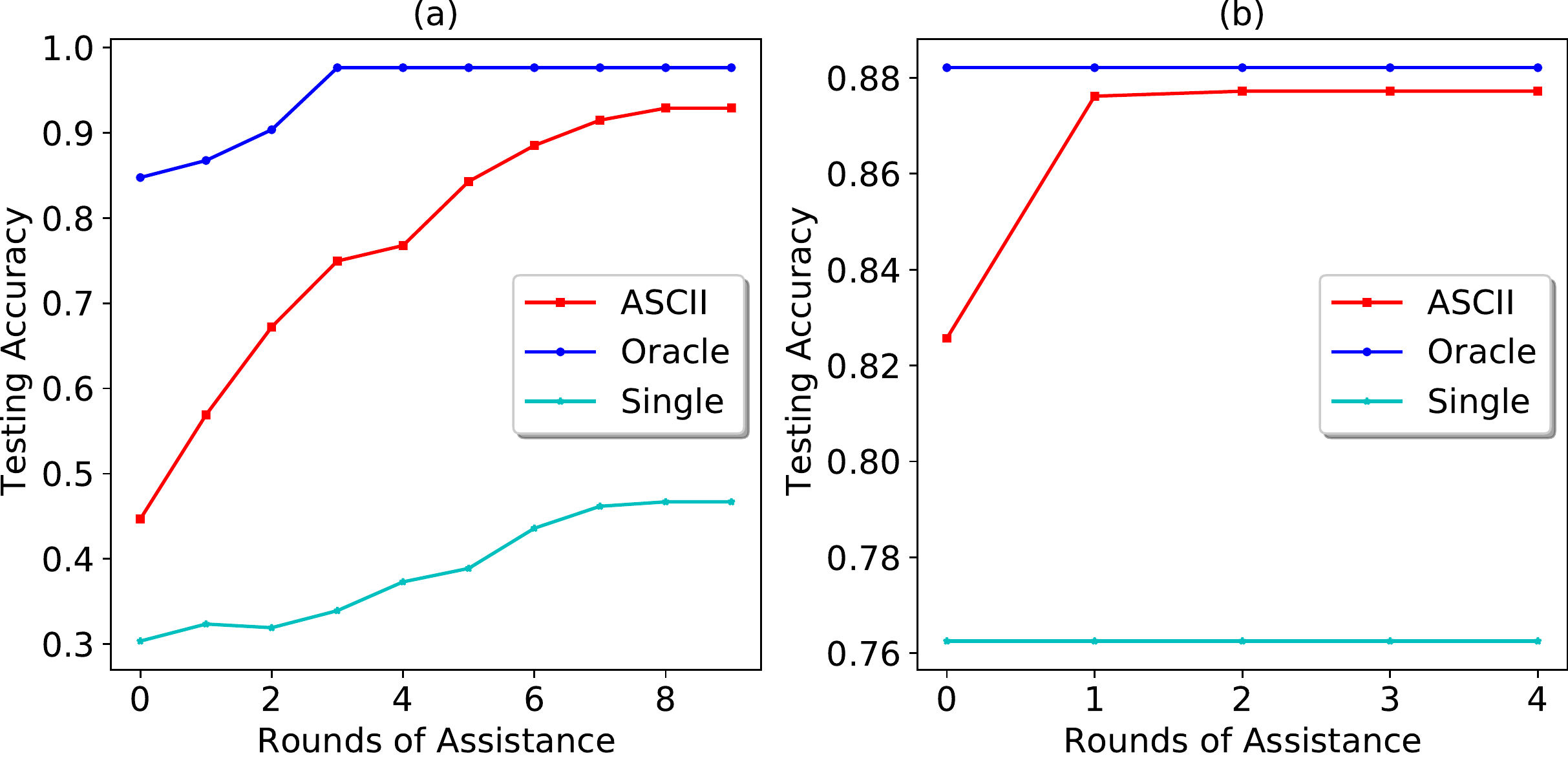}
     \vspace{-0.0in}
     \caption{Out-sample predictive accuracy  for the datasets of (a) Gaussian Blob data (b) Fashion-MNIST, where the transmission costs are improved by around 10 and 195 times respectively, when compared to the oracle approach. The costs are evaluated using the number of bits needed for transmission at $90\%$-oracle test accuracy.
     The methods used on blob and Fashion-MNIST data are random forest and 3-layer neural network, respectively.} % The labels and legends follow the same as Figure~\ref{experiment1}.}
     \label{experiment2}
     \vspace{-0.0in}
  \end{figure}
  
% plt.plot(grid[restrict], y_ip[restrict], 'b', label=r'$X \mid \Delta=0,\Gamma=0, U=-1, V=0$')

%\vspace{-0.2cm}
\subsection{ASCII for reducing transmission cost}
%\vspace{-0.2cm}

We use two examples to illustrate that ASCII can significantly reduce the transmission cost while maintaining  near-oracle performance. Compared with transmitting raw data from $\B$ to $\A$, ASCII only requires transmitting ignorance score (length $n$) and model weight (scalar) at each round of iteration. We use our algorithm to show that ASCII can approximate the oracle, with much less transmission costs when compared with transmitting data from agent $\B$ to  $\A$.

The first is Gaussian Blob data, which is generated from 5 features and 10 classes. Additionally, 
195 redundant features are generated and appended to the above 5 features. We randomly divide these 200 features into 2 agents, each agent holding 100 features. Each agent performs a random forest classifier with the same depth and number of trees.
The result is shown in Figure~\ref{experiment2}.

%The transmission cost from ASCII and transmitting data is 24114.4 and 280000, respectively.
%  \begin{figure}[!htb]
%  \minipage{0.32\textwidth}
%   \includegraphics[width=\linewidth]{figures/blob2learner.pdf}
%   \caption{Blob data: 15 classes, 4 agents}\label{fig_1a}
% \endminipage\hfill
% \minipage{0.32\textwidth}
%   \includegraphics[width=\linewidth]{figures/noisy.pdf}
%   \caption{Blob data:10 classes, 2 agents, 195 noisy features}\label{trans1}
% \endminipage\hfill
% \minipage{0.32\textwidth}%
%   \includegraphics[width=\linewidth]{figures/multimodel.pdf}
%   \caption{Different models in two agents}\label{multimodel}
% \endminipage
% \end{figure}
\begin{figure*}
    \centering
    \includegraphics[width=0.8\linewidth]{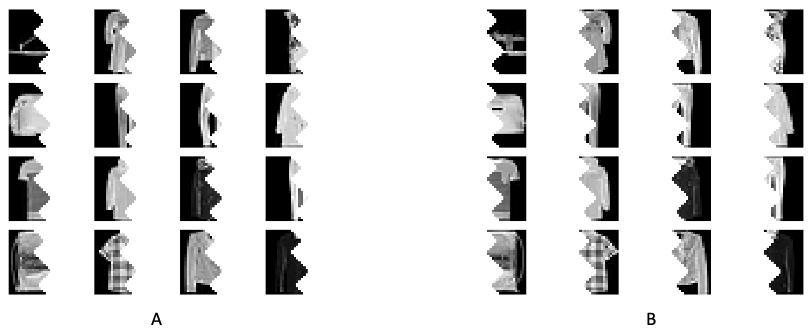}
    \caption{An example pictures in A and B, A holds half of the picture and B holds another half.}
    \label{fig:mnist example}
\end{figure*}
The second example utilizes Fashion-MNIST, an accessible classification dataset with 10 classes in total, each sample being a $28 \times 28$  matrix with entries $0$-$255$. It has $6\cdot 10^4$ training and $10^4$ testing samples, respectively. Suppose that agent $\A$ and $\B$ both hold half part of each image, as illustrated in Figure~\ref{fig:mnist example}. $\B$ does not want to share the picture with $\A$ (possibly due to privacy and transmission costs). 
% We randomly zero half pixels of each picture  and denote the set of   those pictures as agent $\A$.  In contrast, agent $\B$ holds the complementary  information that was erased from agent $\A$. 
Thus, each agent possesses partial information of the same objects.  
We used 3-layer neural networks to build the model, and summarize the results in Figure~\ref{experiment2}. Though neural networks are expressive, a single learner's predictive performance is still restricted because of limited information. The plot shows that  after only two iterations, the testing performance is close to the oracle. As a result, the transmission cost of ASCII compared with the oracle (passing $\B$’s data to $\A$) is reduced over 100 times in terms of the number of transmitted bits. %because ASCII only transmits weight in each iteration. 

The result summarized in Figure~\ref{experiment2} shows that the transmission cost of ASCII compared with the oracle (passing $\B$'s data to $\A$) is reduced over 100 times in terms of the number of transmitted bits.

%%%%  New subsection    %%%%%
%\subsection{ASCII for outperforming other methods}
\subsection{ASCII compared with other methods}
\label{sec_exp:compare method}
In this subsection, we illustrate that ASCII can outperform the variants mentioned in Section~\ref{sec_other}. %We mainly compare ASCII with two other similar methods: 

\begin{figure}
    \centering
    \includegraphics[width=1\linewidth]{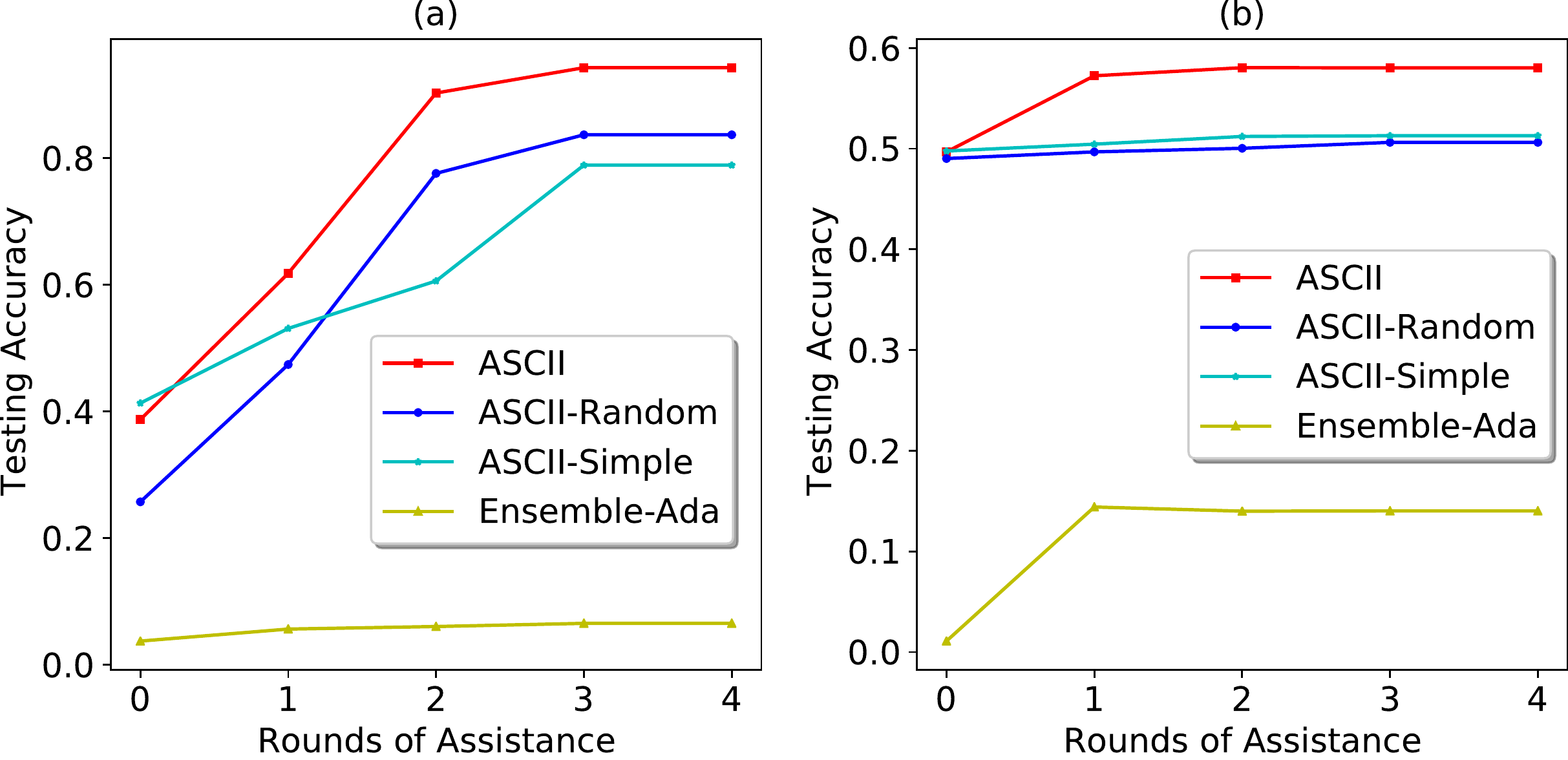}
    \caption{Out-sample predictive accuracy of the proposed method (`ASCII'), the ASCII method in a random assistance manner (`ASCII-Random'),  the ASCII method that only transfer ignorance score (`ASCII-Simple') and the non-assistance Ensemble Adaboost (`Ensemble-Ada')
    against the number of rounds  for the datasets of (a) Blob,  (b) Wine, as described in the text. The model used by each agent in (1)and (2) are   logistic regression and decision tree. Standard errors over 20 independent replications are within 0.04.}
    \label{fig:exp3:multplemethods}
\end{figure}

\textbf{Blob Data} (Synthetic). We generate $20$-class isotropic Gaussian blobs for clustering. For training data, the centralized feature matrix $\bm X \in \mathbb R^{1000\times 20}$ is held by $20$ agents, each holding $1$ heterogeneous feature.
Let $\bm c \in \mathbb R^{1000}$ be the $20$-class label.  Suppose that each agent uses logistic regression for classification. The result is shown in Figure~\ref{fig:exp3:multplemethods}.

\textbf{Red wine quality data}. The red wine classification data set has 1600 data, 11 attributes, and 6 class labels. Suppose that each agent holds a unique feature, and uses a decision tree classifier. 

The result indicates that ASCII can outperform other related methods since ASCII-Simple only considers the interchange of data-level side information. ASCII-Random performs similar or slightly better than ASCII-Simple, but still not as good as ASCII, the possible reason is that ASCII-Random takes the model-level side information interchange into account, but ASCII-Random may result some extreme cases, such as
$\A$ may still interchange with $\A$ at the end of one iteration and the begin of next iteration. This may lose some power of interchanging information, because $\A$ usually can't provide more information other than models. Ensemble Adaboost is the worst since it neither considers model-level or data-level side information, there exists no assistance between agents.
The results are aligned with our explanations in Section~\ref{sec_other},
and it turns out that the transmission of model weight and the ignorance score are both necessary since agents provide their supplement information on samples and models to agent $1$.

%%%%%%%%%%%%%%%%%%%%%%%%%%%%%%%%%%%%%%%
%\vspace{-0.3cm}
\section{Conclusion} \label{sec_con}
%\vspace{-0.2cm}

 In this paper, we proposed a general method for an agent to improve its classification performance by  iteratively interchanging ignorance scores with other agents. %, where the values encode the data regime that needs further cooperation. 
 Our method is naturally suitable for autonomous learning scenarios where private raw data cannot be shared. 
 Moreover, the proposed method allows agents to use private local models or algorithms, which is appealing in many application domains. % since  model-privacy may also be a matter of concern during cooperative learning.

Some future directions are summarized as follows. First, our work addressed classification, and we believe that similar techniques can be emulated to study regression problems. 
%A recent research \cite{xian2020assisted} describes the regression case when agents interchange with each other via residuals. 
Second, from various experiments, we observed that a single agent often achieves near-oracle performance by interchanging with other agents in random orders. This motivates the problem to study the most efficient order of interchanging information to attain the optimum. Another open problem is to study how asynchronous interchange, meaning different orders at two rounds, will influence the learning efficiency. 
\balance 
%\nocite{*} % to test all bib entrys
\bibliographystyle{IEEEtran}
%\bibliography{reference.bib}

\ifCLASSOPTIONcaptionsoff
  \newpage
\fi

\end{document}